# EarthGAN: Can we visualize the Earth's mantle convection using a surrogate model?

Tim von Hahn*, Chris K. Mechefske

Queen's University, Kingston, Canada

**ABSTRACT**

Scientific simulations are often used to gain insight into foundational questions. However, many potentially useful simulation results are difficult to visualize without powerful computers. In this research, we seek to build a surrogate model, using a generative adversarial network, to allow for the visualization of the Earth's Mantle Convection data set on readily accessible hardware. We present our preliminary method and results, and all code is made publicly available. The preliminary results show that a surrogate model of the Earth's Mantle Convection data set can generate useful results. A comparison to the "ground-truth" model is provided.

## 1 INTRODUCTION

Modern high-performance computers, coupled with specific domain knowledge, can create realistic simulations. This is the case with the Earth's Mantle Convection simulation [1]. However, the results from these large-scale simulations, based on detailed scientific formulations, are difficult to explore without specialized (and expensive) hardware.

Our central hypothesis is that we can create a surrogate model of the Earth's Mantle Convection data set and leverage it for real-time visualization. The final surrogate model must run locally (no off-premises computing), in a web browser, and be of similar visual fidelity to the original data set.

The following section describes a surrogate model and why the problem is of interest. We then introduce the method we are pursuing using a state-of-the-art generative adversarial network (GAN). Finally, we discuss preliminary results and future work.

### 1.1 What is a surrogate model?

"A surrogate model is an approximation of the input-output data obtained by [a] simulation." [2] The process of constructing surrogate models can be achieved through traditional machine learning techniques, or through deep learning.

We posit that the Earth Mantle Convection data set has strong "pre-defined regularities arising from the underlying low-dimensionality and structure of the physical world" [3] that make it amenable for use in surrogate modeling. In particular, we believe that the large size of the data set, coupled with advances in deep learning, can produce a high-fidelity surrogate model.

### 1.2 Why build a surrogate model?

The ability to visualize the results of a large scientific simulation, such as the Earth's Mantle Convection data set, is prohibitive without costly computational infrastructure. Therefore, the public, and researchers, without access to high performance computing, cannot easily visualize and explore the results of these simulations.

---

* 18tcvh@queensu.ca

Ideally, a surrogate model, once trained, would be inexpensive to query. Therefore, the surrogate model could be used to generate realistic samples on accessible computer hardware. With this logic, we aim to train a surrogate model to replicate the Earth's Mantle Convection data set, and then use the model for data visualization. The "cost" of visualizing the data is then paid for once, upfront, during the training of the surrogate model.

Building a surrogate model of a complex scientific data set is a significant challenge. Our preliminary method, described below, uses the tools of deep learning, and specifically the use of a generative adversarial network (GAN). GANs consist of paired generator and discriminator neural networks [4]. The generator creates "fake" outputs, in our case a fake sample of the Earth's Mantle Convection data, and the discriminator attempts to discern (discriminate) between the "fake" output and the ground-truth data.

The tools and methods employed in deep learning can be opaque – it is as much an "art" as a science. Even more so, GANs are recognized as difficult to train. Through open sourcing our research, and engaging with the community, we endeavor to accelerate the learning process for others and provide a foundation for further research. All research and code are available on GitHub (github.com/tvhahn/EarthGAN).

## 2 CURRENT APPROACH

### 2.1 Model Architecture

The model architecture, and the model implementation, are largely derived from the work of Li et al. [5]. They trained a GAN to create super-resolution cosmological simulations based on low-resolution input. Both the input and output from their GAN are 3D volumes, and thus, transferrable to the Earth's Mantle Convection data set. The architecture they use leverages techniques from StyleGAN2 [6], the Wasserstein GAN [7], and the Wasserstein GAN with a gradient penalty [8].

Importantly, Li et al. make use of a conditional-GAN (cGAN) [9] in their work. This method is also used in our current approach. Traditionally, a GAN uses random noise as an input to the generator. With a cGAN, the input to the generator, and the discriminator (sometimes called a critic), is conditioned with additional information. In our case, the additional information, as input to the generator, is a downscaled representation of the Earth's Mantle Convection data. The same downscaled version is then upscaled, through linear interpolation, and used as additional input to the discriminator. The use of the cGAN makes the approach partially deterministic.

### 2.2 Data Preparation

The process of data preparation is illustrated in Figure 2 in the appendix. The original data, regardless of variable or radial dimension, has a height and width of 180x360, representing the latitudinal and longitudinal dimensions of the Earth, respectively. From the outset, the data was scaled down by 40% to achieve a height and width of 108x216. This became the ground truth data used in the discriminator.

From there, the data was further down sampled by a factor of 8. Because the height dimension of 108 is not fully divisible by 8, the top and bottom dimensions were padded, through a mirror, by 2 each. The final width and height of the downscaled data was thus 14x27.

The generator takes the input data, and through a series of convolutions, projections, and interpolations, as described in StyleGAN2, produces an output; that is, the "fake" data. The output from the generator will be of the same scale as the ground-truth but cropped due to the operations of the generator. Therefore, to achieve a properly cropped output from the generator to match the ground-truth, the input was additionally padded, again by mirroring, along the latitudinal axis.

The final height and width dimensions of the input, to the generator, was 20x10. The smaller width dimension, of 10, was selected to reduce the memory requirements during the training. Finally, the output from the generator was 118x38 (height and width), representing a longitudinal slice of approximately 63 degrees. Both the input data and the ground-truth data were scaled between 0 and 1.

Only 30 radial layers, equally spaced between the outermost and innermost radial layers, were included in the input to the generator. In addition, only the temperature, and three velocity vector layers (vx, vy, and vz) were used. Abnormalities in the other variables were consistently found, and thus, were omitted at this preliminary stage. As such, the final size of the input data was of a shape (4, 30, 20, 10), representing the variable, radial dimension, height (latitude), and width (longitude), respectively. Likewise, the output from the generator was (4, 198, 118, 38).

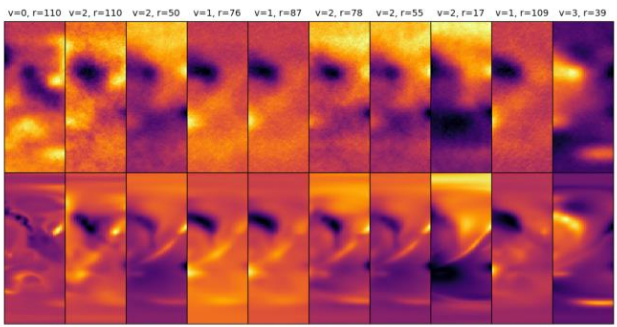

Figure 1: Top row shows "fake" outputs from the generator, and the bottom row is the corresponding ground-truth. *v* represents the variable (temperature, vx, vy, and vz, respectively). *r* represents the radial layer, where *r* = 0 is closest to the Earth's core.

## 2.3 Training

Training took place over several days on a single P100 GPU and consisted of 46 epochs. The data set was significantly augmented by rotating each sample by a random amount about the Earth's axis. Unfortunately, the training consistently encountered the 12 GB GPU memory constraint. As such, only a mini-batch size of one was used. Consequently, the training was slow, which must be addressed in further iterations of the experiment. In addition, the maximum channel size in the convolutions (within the generator and discriminator) was capped at 128.

## 3 PRELIMINARY RESULTS AND FUTURE WORK

The preliminary results demonstrate that the model is training. However, fine features cannot yet be generated. Figure 1 shows a random selection of "fake" results and their corresponding ground-truth. Further examples are in the appendix.

The current size of the generator, when serialized, is 2.7 MB, and a low-end GPU (4GB NVIDIA 980m) can readily generate the fake Earth Mantle Convection samples. The ground-truth data, for context, has a size of 18.5 GB, uncompressed. This difference in size, as illustrated in Figure 3 in the appendix, demonstrates the promise of a surrogate model. However, significant additional work is required before firm conclusions on the effectiveness of the method can be made.

In the short term, we wish to speed up the training. We will train only on one variable (temperature), introduce multi-GPU training, and use mixed precision training. Also, we will increase the maximum channel size in the convolutions from 128 to 256.

Longer term, we believe that we must enhance the informational quality of the conditional input to the discriminator. Li et al. [5] note that the enhanced quality of their conditional input was key in achieving sharp features in their super-resolution cosmological simulations. They used a "differentiable CIC operation", and likewise, there may be similar abstractions that can be leveraged for the Earth's Mantle Convection data set. Geological domain expertise may be of benefit. Otherwise, we may attempt automatic segmentation with a U-Net [10].

## 4 CONCLUSION

Can a surrogate model of the Earth's Mantle Convection data set be built such that it can readily run in a web-browser and produce high-fidelity results? The preliminary results are encouraging, but much additional work is required to fully answer this question. Ultimately, a surrogate model of the Earth's Mantle Convection data set, and other scientific data sets, could be used to perform data exploration without expensive hardware. We trust that others will find this preliminary work useful and use it to further advance the field.

**APPENDIX**

Figure 2, below, illustrates the data preparation steps. Both the ground-truth data, and the input to the generator, are scaled between 0 and 1 before use in

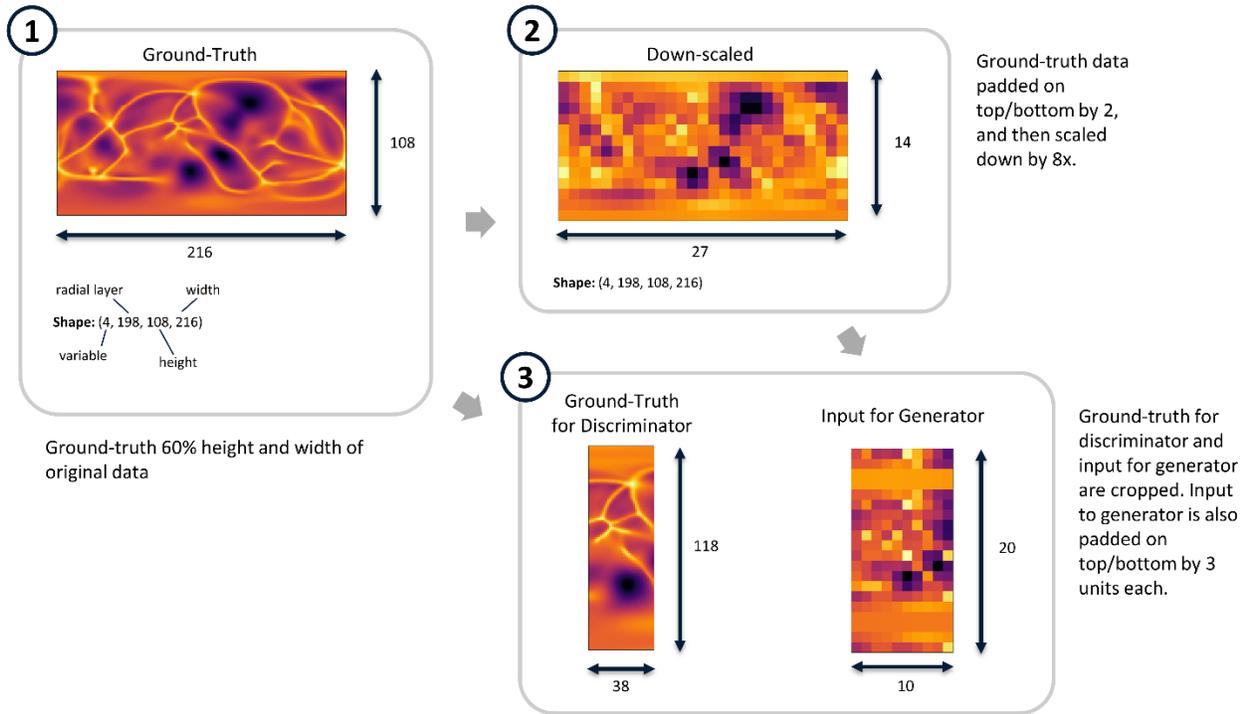

Figure 2: The process of preparing the data. 1) The ground-truth data is scaled to 60% of original size. 2) The data is down-scaled by 8x. 3) The ground truth, used in the discriminator, is cropped. The input the generator is also cropped.

Figure 3, below, compares the size of the original ground-truth data (as used in this experiment) to the size of the surrogate model and associated generator input. We expect the model size to increase as the experiment progresses and we increase the max channel count from 128 to 256. However, additional tools, such as mixed precision, can be used to reduce the model and model input data size.

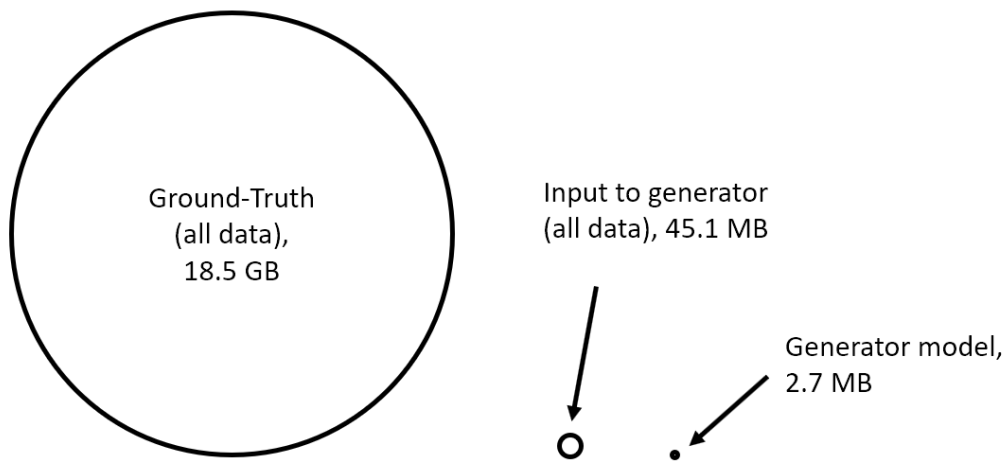

Figure 3: The size of the ground-truth data, used to train the generator model, is much larger than the size of the input data and generator model. The size difference illustrates, on one dimension, how a surrogate model can be of benefit.

Below are several examples of the model's performance, taken during training.

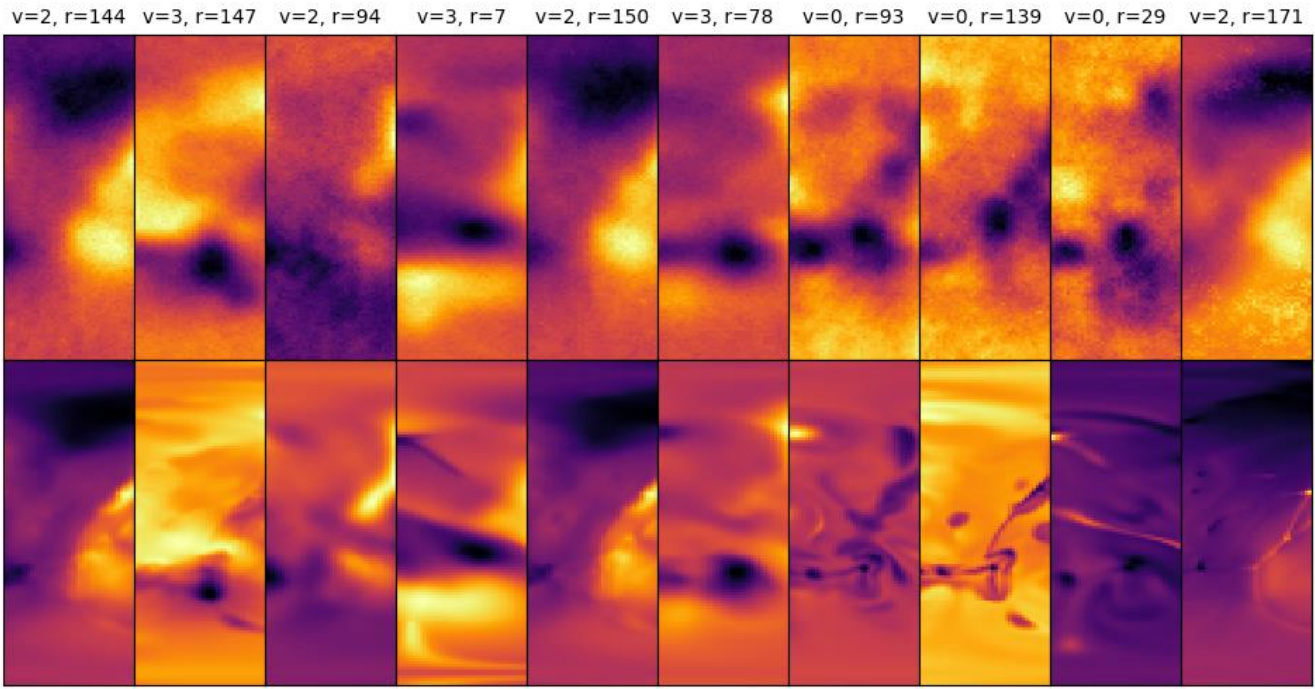

Figure 4: Random samples of ground-truth (bottom row) and the corresponding output from generator (top). From epoch 28. *v* represents the variable (temperature, vx, vy, and vz, respectively). *r* represents the radial layer, where *r* = 0 is closest to the Earth's core.

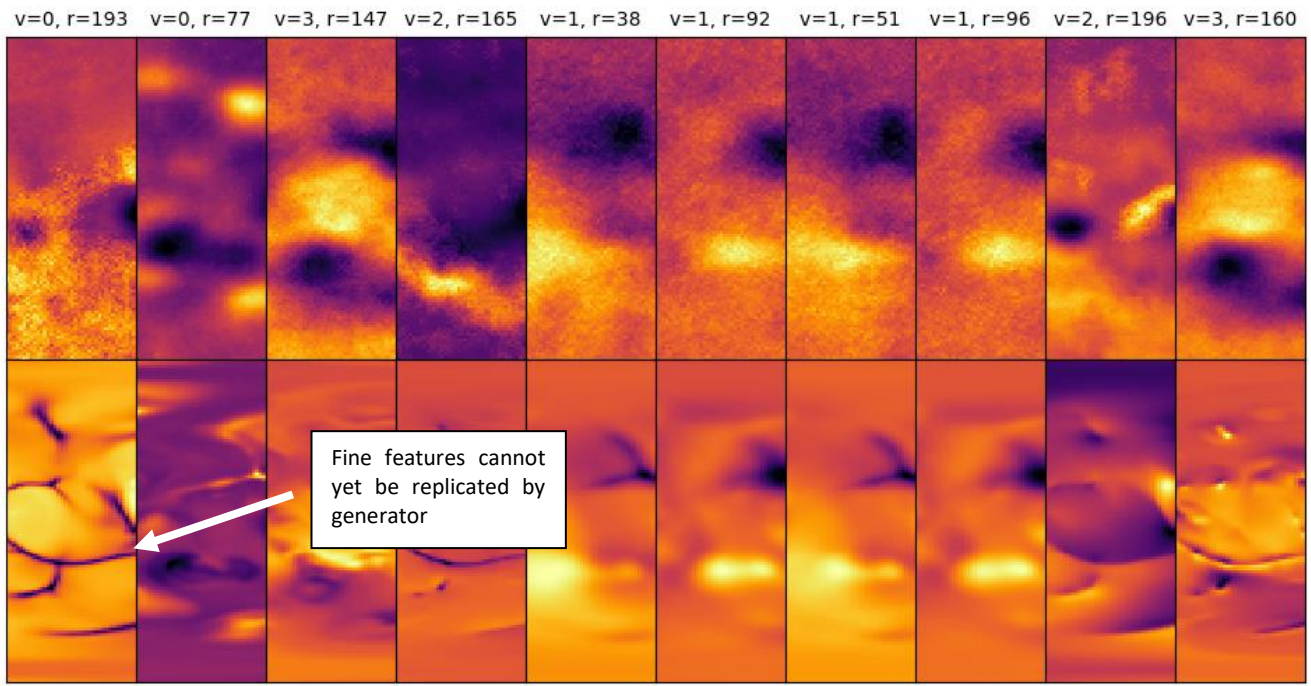

Figure 5: From epoch 28. Fine features cannot yet be replicated well by the generator.

Occasionally, during training, the generator "fake" output produces vertical lines, as shown in Figure 6, below. We believe these vertical lines are the generator attempting to replicate the fine features. This is a curious result and we are interested in understanding why this happens.

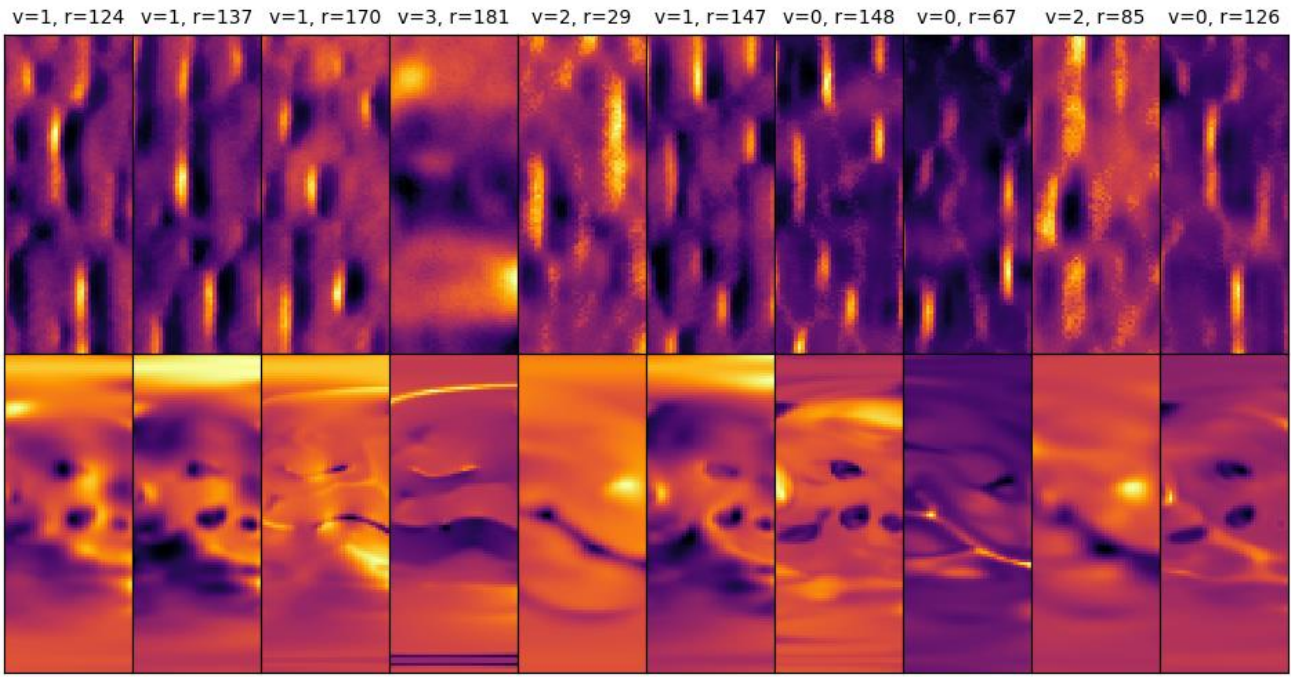

Figure 6: From epoch 16. The top row of "fake" outputs from the generator show a interesting pattern of vertical lines.

The output from the generator is only a "slice" of the Earth's mantle. Slices can be concatenated together to produce a full representation of the Earth's Mantle.

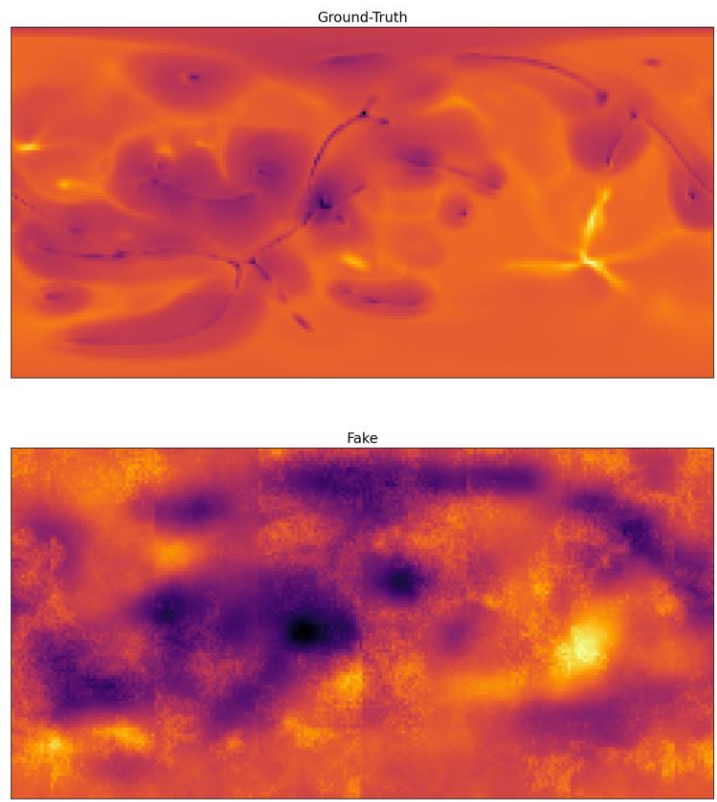

Figure 7: Epoch 36, timestep 1, radial index 165.

The partially stochastic nature of the GAN (through noise injection) produces perceptible lines in the concatenated outputs, as shown below. This may be addressed by averaging several outputs together. However, the issue may resolve itself as the model results improve.

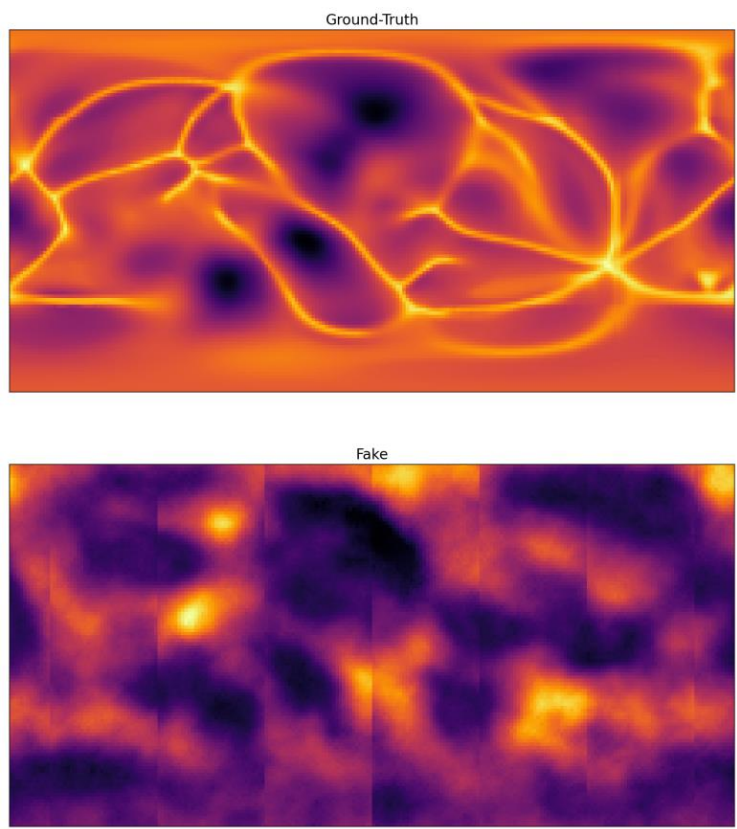

Figure 8: Epoch 36, timestep 1, radial index 0

We are pleased with these preliminary results so far and are excited to improve the method. Much work is still required. Feel free to contribute, fork the code, or provide suggestions at https://github.com/tvhahn/EarthGAN.